\documentclass[a4paper]{article}
\usepackage{makeidx}  

\usepackage{graphicx,subfigure}
\usepackage[export]{adjustbox}
\usepackage{amsmath} 
\usepackage{amssymb} 

\usepackage{lineno}
\usepackage{multicol} 
\usepackage{tabularx}
\usepackage[export]{adjustbox}
\usepackage{multirow}
\usepackage{hhline}
\usepackage{authblk}

\newcolumntype{P}[1]{>{\centering\arraybackslash}p{#1}}
\newcolumntype{M}[1]{>{\centering\arraybackslash}m{#1}}

\begin{document}

%
%

\title{Aligning Manifolds of Double Pendulum Dynamics Under the Influence of Noise \thanks{The final version will appear on the proceedings of the 25th International Conference on Neural Information Processing (ICONIP 2018).}}

\author{Fayeem Aziz\thanks{MdFayeemBin.Aziz@uon.edu.au}}
\author{Aaron S.W. Wong}
\author{James S. Welsh}
\author{Stephan K. Chalup\thanks{Stephan.Chalup@newcastle.edu.au}}

\affil{School of Electrical Engineering and Computing \\  The University of Newcastle, \\ Callaghan NSW 2308, Australia}

\maketitle


\begin{abstract}
This study presents the results of a series of simulation experiments that evaluate and compare four different manifold alignment methods under the influence of noise. The data was created by simulating the dynamics of two slightly different double pendulums in three-dimensional space. The method of semi-supervised feature-level manifold alignment using global distance resulted in the most convincing visualisations. However, the semi-supervised feature-level local alignment methods resulted in smaller alignment errors. These local alignment methods were also more robust to noise and faster than the other methods.  
\end{abstract}
\section{Introduction}
Manifold alignment of two data sets assumes that they have similar underlying manifolds. The aim is to find a mapping between the two data sets so that the underlying manifold structure can be better recognised. Due to its generality manifold alignment has great potential to be useful in various domains. In the past it has been applied to facial expression analysis by image sequence alignment~\cite{cui:2012,Ham:2005,pei:2012,Xianwang:2010,zhai:2010}, graph matching~\cite{Escolano:2011}, image classification~\cite{guerrero:2014,Hsiuhan:2011}, and bioinformatics~\cite{Wang:2008}.  However, due to its high computational demands and the complexity of manifold data, manifold alignment still requires substantial further research and development to increase its impact in practical applications.   

Real-world data is often sampled from a high-dimensional space and can be modelled as a set of points that lie on a low-dimensional non-linear manifold~\cite{Huang:2009}. This also applies to data collected from robot kinematics or human motion which could be represented in form of dynamical system trajectories on low-dimensional manifolds~\cite{Chalodhorn:2007}. It has been shown that motion data can be transferred from a human to a robot or between robots by aligning the corresponding manifolds~\cite{Bocsi:2013,Chalodhorn:2010}. In this context, the ability of manifold learning and manifold alignment to represent non-linear data is critical~\cite{TenenBaum:2000}. An example that can illustrate this is a double pendulum and its non-linear dynamics~\cite{Jensen:1998}. 

The contribution of this study is an experimental comparison of several existing manifold alignment algorithms using data sampled from the motions of two simulated double pendulums in three-dimensions. Previous restricted pilot experiments, had focused on a pendulum in two dimensions~\cite{fayeem:inpress}. The present paper also it investigates the stability of the methods under the influence of noise. It introduces the concepts of normalised distances along with pairwise correspondence measures~\cite{fan:2016}. The experimental evaluation focuses on visualisations and measuring the proximity of alignments and the execution time of the algorithms.
\begin{figure}[!ht]
  \centering     
  \leavevmode
  \includegraphics[width=0.37\textwidth]{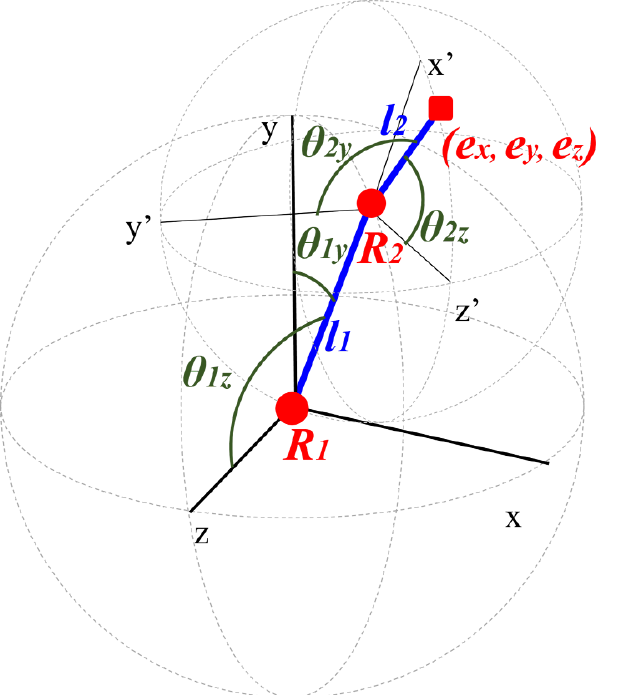}
  \caption{Double pendulum motion in  3D. $x$, $y$ and $z$ are the local axes of limb $l_1$ and $x'$, $y'$ and $z'$ are the local axes of limb $l_2$. $\left(e_x,e_y,e_z\right)$ is the end-effector.}
  \label{fig:1}
\end{figure}
\section{Double Pendulum Data}\label{sec:pend-data}
The simulated data used in our experiments represents the motion of a three-dimensional double pendulum similar to a two-limb robot arm as shown in Figure~\ref{fig:1}. The two limbs are denoted $l_1$ and $l_2$ and the joints are denoted $R_1$ and $R_2$. One end of $l_1$ is fixed at $R_1$ and can rotate around this joint. The other end of $l_1$ is attached to $l_2$ at joint $R_2$. $l_2$ can rotate freely around joint $R_2$. The end point of the arm, that is, the end of $l_2$ not attached to $R_2$, is also known as the end-effector and has coordinates $\left(e_x,e_y,e_z\right)$. The coordinate system follows the right-hand rule and has its origin at joint $R_1$. $\theta_{1y}$ and $\theta_{1z}$ are the angles of $l_1$ with the $y$ and $z$ axes, respectively, and $\theta_{2y'}$ and $\theta_{2z'}$ are the angles of $l_2$ with the $y'$ and $z'$ axes, respectively. The feature vector for each sample point was calculated from the kinematics at the joints and the end-effector coordinates were calculated using forward kinematics:\begin{equation}
(e_x, e_y, e_z,  \cos\theta_{1y}, \cos\theta_{1z}, \cos\theta_{2y'},  \cos\theta_{2z'}, \sin\theta_{1z}, \sin\theta_{1y},  \sin\theta_{2y'}, \sin\theta_{2z'})
\end{equation}
The data sets were generated from two similar double pendulums, which had different limb lengths and slightly different limb length ratios, where we restricted the experiments to the case $l_2 < l_1$ (Figure~\ref{fig:1}):
\begin{itemize}
\item[]Pendulum 1: $(l_2/l_1) = 0.75/1.25 = 0.60$
\item[]Pendulum 2: $(l_2/l_1) = 1.25/1.56 = 0.80$,
\end{itemize}
The data was acquired using increments of $30^\circ$ on all four axes $y$, $z$, $y'$ and $z'$. As a result, the number of instances was $(360/30)^4 = 20736$ and the size of each of these data sets was $20736\times11$. If the following sections refer to data sets $X$ and $Y$ it is assumed that the data is arranged in the form of matrices of dimension $20736\times11$. The correspondence subsets of $X$ and $Y$ comprise corresponding points in both sets  that have the same or  similar joint angles. They were selected by using $90^\circ$ joint angle steps between two instances in the same data set. Therefore each of the two correspondence subsets had $(360/90)^4 = 256$ instances which was about $10\%$ of the data set. 

In the experiments, uniformly distributed white noise was added in two different ways to the data. First, noise was added to the joint angles and the noise range was incremented from $0^\circ$ to $\pm10^\circ$ in steps of $1^\circ$. The second type of noise was added to the end effector coordinates. The range of this coordinate noise was increased from $0$ to $\pm1.0$ in steps of $0.1$. 
\begin{figure}[!ht]
  \centering     
  \leavevmode
  \subfigure[Approach 1]{\includegraphics[width=0.45\textwidth,height = 4cm,frame]{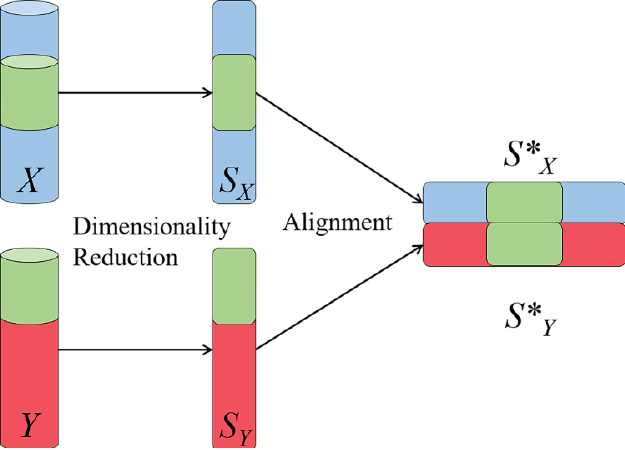}} 
  \subfigure[Approach 2]{\includegraphics[width=0.45\textwidth,height = 4cm,frame]{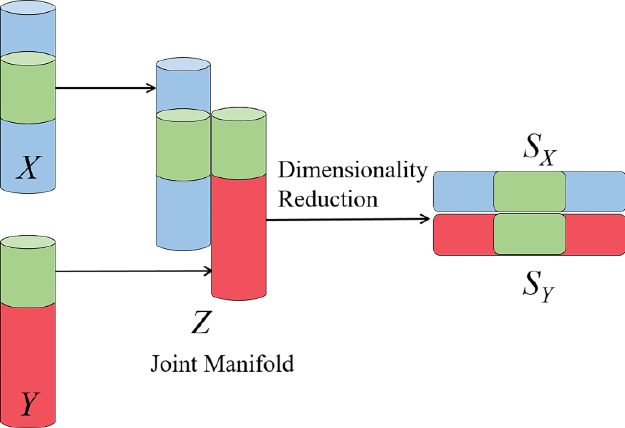}} 
  \caption{(a)~Approach 1 : Dimensionality reduction is applied to $X$ and $Y$, which yields low-dimensional manifolds $S_X$ and $S_Y$, respectively. The manifolds are then mapped into a joint space as aligned manifolds $S^*_X$ and $S^*_Y$. (b)~Approach 2: A joint matrix $Z$ is generated from $X$ and $Y$ by regularisation of correspondence information. Then, dimensionality reduction is applied to obtain the low-dimensional aligned manifolds $S_X$ and $S_Y$. The green areas in both approaches indicate the correspondence subsets.}
  \label{fig:2}
\end{figure}
\section{Manifold Alignment Methods}
We address two general approaches to align manifolds. Each of them has two stages (Figure~\ref{fig:2}). In Approach 1 the data sets are first mapped into a low-dimensional space and then alignment is performed as described by Wang et al.~\cite{Wang:2008}. In Approach 2, first a joint manifold is created to represent the union of the given manifolds and then the joint manifold is mapped to a lower dimensional latent space as described by Ham et al.~\cite{Ham:2005}. Approach 2 was also described as a generalised semi-supervised manifold alignment framework by Wang and Mahadevan~\cite{Wang:2009a}, where the joint manifolds can be constructed using different characteristics of the data based on the application.

Wang and Mahadevan~\cite{Wang:2009a} further distinguished between two levels of manifold alignment: instance-level alignment (Inst) and feature-level alignment (Feat). For instance-level alignment, their method computes a non-linear low-dimensional embedding based on an alignment of the instances in the data. The embedding results in a direct matching of corresponding instances in alignment space. In contrast, feature-level alignment builds on mapping functions of features which map any associated instance or set of instances into the newly aligned domain.

\begin{figure}[!ht]
  \centering     
  \leavevmode
  \includegraphics[width=\textwidth]{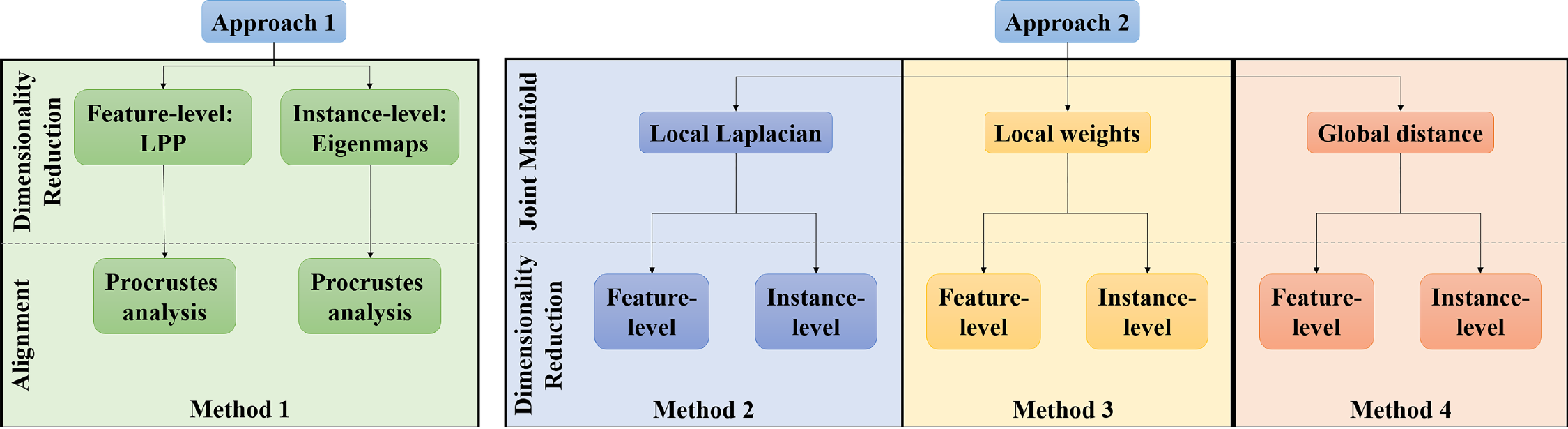}
  \caption{Overview of methods used in the experiments where the general structure of Approach 1 and 2 is detailed in Figure~\ref{fig:2}.}
  \label{fig:3}
\end{figure}

The present study applied the following four methods to the double pendulum data sets. Method 1 followed Approach 1 (\cite{Wang:2008} and Figure~\ref{fig:2}~(a).) and methods 2-3 followed the semi-supervised Approach 2 (\cite{Wang:2009a} and Figure~\ref{fig:2}~(b)). Further, each of the four methods had an instance-level and a feature-level version that was addressed in separate experiments (see overview in Figure~\ref{fig:3}):
\begin{itemize}
\item {\bf Method 1:} Locality Preserving Projection (LPP)~\cite{Niyogi:2004} is used for feature-level dimensionality reduction, and Laplacian eigenmaps (eigenmaps)~\cite{Belkin:2003} is used for instance-level dimensionality reduction, as the first stage, in two different experiments. In the second stage (Figure~\ref{fig:2}~(a)) we followed~\cite{Wang:2008} and employed Procrustes analysis~\cite{Luo:1999} to align the two data sets in low-dimensional space. 
\item {\bf Method 2:} For semi-supervised manifold alignment preserving local geometry~\cite{Wang:2009a}, first, the joint manifold $Z$ was calculated using the graph Laplacians for $X$ and $Y$. Then, eigenvalue decomposition of the joint manifold provided instance-level alignment and generalised eigenvalue decomposition provided feature-level alignment.
\item {\bf Method 3:} For semi-supervised manifold alignment using local weights~\cite{Wang:2011a} the intermediary joint manifold $Z$ was formed by weights for $X$ and $Y$, which were calculated for each set using a $k$-nearest neighbour graph and the heat kernel. Similar to Method 2 the low-dimensional manifolds were mapped by eigenvalue decomposition of joint matrices in two experimental components. 
\item {\bf Method 4:} For semi-supervised manifold alignment preserving global geometry~\cite{Wang:2013} the joint manifold $Z$ was generated using the global distances of corresponding pairs in $X \cup Y$. The eigenvalue decomposition of $Z$ provided the dimensionality reduction to obtain the aligned low-dimensional manifolds in the case of instance-level alignment. Generalised eigenvalue decomposition was used to reduce the dimensionality in the case of feature-level alignment.
\end{itemize}

\section{Performance Evaluation}\label{sec:norm}
To measure the performance of the alignment methods, we assumed that all instances of one data set can be mapped one-to-one to instances in the other data set. The widths of the resulting manifolds originating from the same data set could vary for the different methods. As a result of scaling, the distances between corresponding points could be larger between ``wider'' manifolds and smaller between ``narrower'' manifolds. To obtain comparable distance measurements between the two aligned low-dimensional manifolds, the distances were normalised by the maximum width of the two manifolds as follows:
\begin{equation} \label{eq:1}
D_{i} = \frac{\|S_X(i)-S_Y(i)\|}{\max_{\substack{j = 1,...,n\\k = 1,...,n}}(\|S_X(j)-S_X(k)\|,\|S_Y(j)-S_Y(k)\|)}
\end{equation} 
where $S_X(i)$ and $S_Y(i)$ are corresponding points for $i=1,...,n$ and $n$ is the number of points that represent each of the manifolds. The matching errors of the alignments were measured using the average of the $D_{i}$, denoted by $\Delta$. That is, $\Delta$ indicates the closeness of the aligned manifolds in low-dimensional space. The standard deviation of the $D_{i}$, denoted by $\sigma$, represents the consistency or smoothness of the alignment where a smaller $\sigma$ indicates a smoother alignment. $\Delta$ and $\sigma$ could be used to measure the quality of an alignment if it was successful.

\section{Results}
\begin{figure}[!ht]
		
\begin{tabularx}{\textwidth}{M{0.1\textwidth}M{0.08\textwidth}M{0.1\textwidth}M{0.08\textwidth}M{0.1\textwidth}M{0.08\textwidth}M{0.1\textwidth}M{0.08\textwidth}}\\
\hline\noalign{\smallskip}
\multicolumn{2}{c}{\multirow{2}{*}{\parbox{2.8cm}{\centering Procrustes analysis}}} 
&
\multicolumn{2}{c}{\multirow{2}{*}{\parbox{2.5cm}{\centering Local Laplacian}}} 
&
\multicolumn{2}{c}{\multirow{2}{*}{\parbox{2.0cm}{\centering Local weight}}} 
& 
\multicolumn{2}{c}{\multirow{2}{*}{\parbox{2.5cm}{\centering Global distance}}} 
\\ & & & & & & & \\
  Feat & Inst 
& Feat & Inst 
& Feat & Inst 
& Feat & Inst 
\\ \hline\vspace{0.15cm}\\

  \includegraphics[width = 0.1\textwidth]{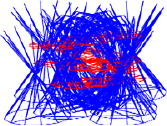}
& \includegraphics[width = 0.08\textwidth]{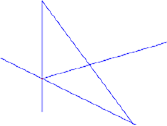} 
& \includegraphics[width = 0.1\textwidth]{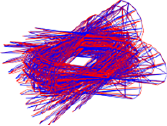} 
& \includegraphics[width = 0.08\textwidth]{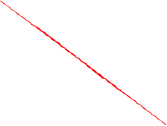}
& \includegraphics[width = 0.1\textwidth]{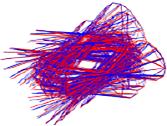}
& \includegraphics[width = 0.08\textwidth]{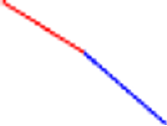}
& \includegraphics[width = 0.1\textwidth]{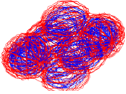}
& \includegraphics[width = 0.08\textwidth]{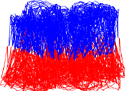}

\\\hline\noalign{\smallskip}

\end{tabularx}

	\vskip -0.5cm
	\caption{3D motion manifold alignments using snapshots in $90^\circ$ steps for one limb and $10^\circ$ steps for the other limb. The output of the feature-level manifold alignment using global distance (column 7) displays six small spheres representing the motion of $l_2$ which are distributed on a larger sphere representing the motion of $l_1$.}
	\label{fig:4} 
\end{figure}

Both limbs of the double pendulum rotated freely in three dimensions. The resulting manifolds were too complex to be visualised in full in a three-dimensional graph. Therefore, we used snapshots of $90^\circ$ steps for the motion of limb $l_1$ and of $10^\circ$ steps for limb $l_2$.  Figure~\ref{fig:4} shows that feature-level manifold alignment using global distance resulted in six small spheres that were distributed regularly on a bigger sphere. The small spheres represent the motion of limb $l_2$ and the bigger sphere represents the motion of limb $l_1$. Ergo this figure can be interpreted as sections in time of the expected shape of the aligned manifolds of pendulum motion. The results from the other three methods for feature level alignment show complex shapes that include partially collapsed submanifolds. For the instance-level cases the manifolds collapsed completely into line segments.

\begin{figure}[!ht]
		\small
\begin{tabularx}{1\textwidth}{M{0.06\textwidth}|M{0.15\textwidth}M{0.15\textwidth}M{0.15\textwidth}M{0.15\textwidth}}
\hline\noalign{\smallskip}
\multirow{2}{*}{\parbox{1.0cm}{\centering Noise\\level}} 
& 
\multicolumn{1}{c}{\multirow{2}{*}{\parbox{2.4cm}{\centering Method 1: Pro-  crustes analysis}}} 
&
\multicolumn{1}{c}{\multirow{2}{*}{\parbox{2.4cm}{\centering Method 2: Local  Laplacian}}} 
&
\multicolumn{1}{c}{\multirow{2}{*}{\parbox{2.4cm}{\centering Method 3: Local weights}}}
&
\multicolumn{1}{c}{\multirow{2}{*}{\parbox{2.4cm}{\centering Method 4: Global distance}}}
\\ & & & & \\
\hline\\
$\pm0^{\circ}$
& \includegraphics[width = 0.15\textwidth]{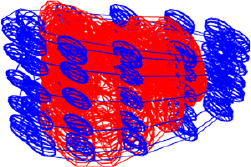}
& \includegraphics[width = 0.15\textwidth]{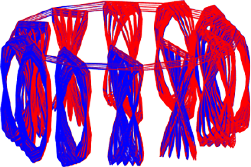} 
& \includegraphics[width = 0.15\textwidth]{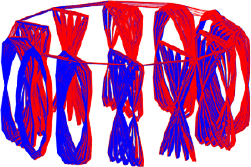}
& \includegraphics[width = 0.15\textwidth]{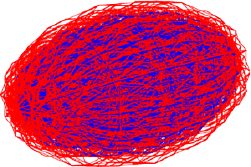}
\\
$\pm1^{\circ}$
& \includegraphics[width = 0.15\textwidth]{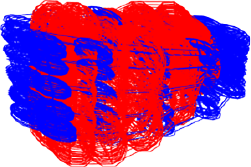}
& \includegraphics[width = 0.15\textwidth]{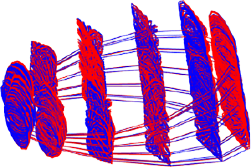} 
& \includegraphics[width = 0.15\textwidth]{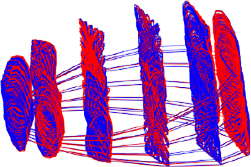}
& \includegraphics[width = 0.15\textwidth]{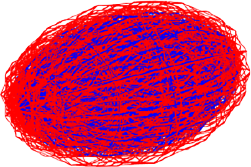}
\\
$\pm2^{\circ}$
& \includegraphics[width = 0.15\textwidth]{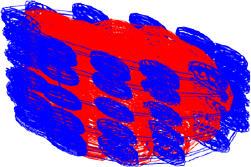}
& \includegraphics[width = 0.15\textwidth]{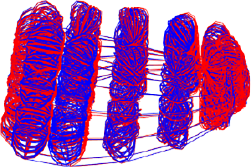} 
& \includegraphics[width = 0.15\textwidth]{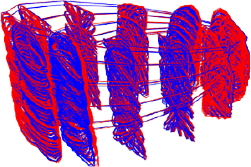}
& \includegraphics[width = 0.15\textwidth]{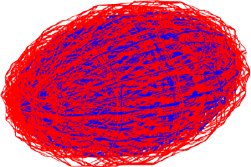}
\\
$\pm3^{\circ}$
& \includegraphics[width = 0.15\textwidth]{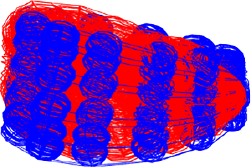}
& \includegraphics[width = 0.15\textwidth]{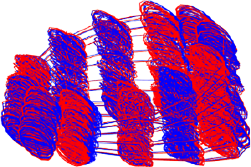} 
& \includegraphics[width = 0.15\textwidth]{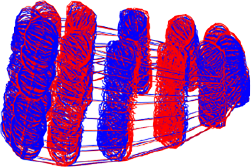}
& \includegraphics[width = 0.15\textwidth]{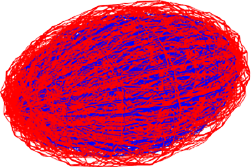}
\\
$\pm4^{\circ}$
& \includegraphics[width = 0.15\textwidth]{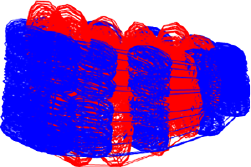}
& \includegraphics[width = 0.15\textwidth]{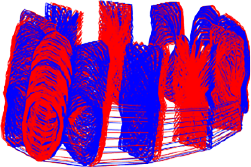} 
& \includegraphics[width = 0.15\textwidth]{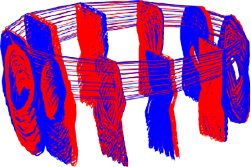}
& \includegraphics[width = 0.15\textwidth]{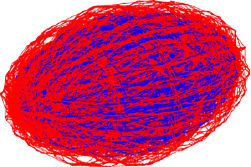}
\\
$\pm5^{\circ}$
& \includegraphics[width = 0.15\textwidth]{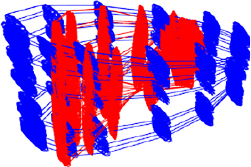}
& \includegraphics[width = 0.15\textwidth]{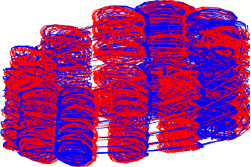} 
& \includegraphics[width = 0.15\textwidth]{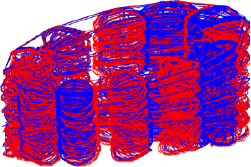}
& \includegraphics[width = 0.15\textwidth]{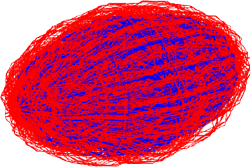}
\\
$\pm6^{\circ}$
& \includegraphics[width = 0.15\textwidth]{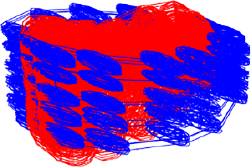}
& \includegraphics[width = 0.15\textwidth]{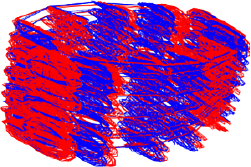} 
& \includegraphics[width = 0.15\textwidth]{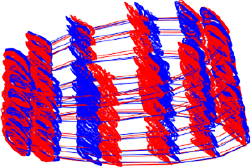}
& \includegraphics[width = 0.15\textwidth]{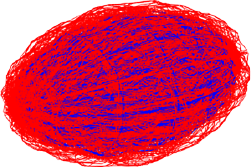}
\\
$\pm7^{\circ}$
& \includegraphics[width = 0.15\textwidth]{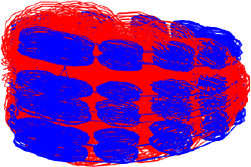}
& \includegraphics[width = 0.15\textwidth]{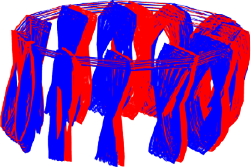} 
& \includegraphics[width = 0.15\textwidth]{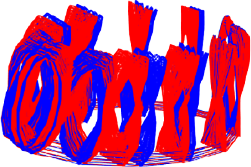}
& \includegraphics[width = 0.15\textwidth]{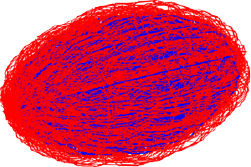}
\\
$\pm8^{\circ}$
& \includegraphics[width = 0.15\textwidth]{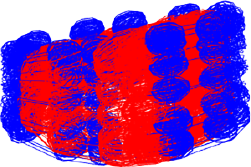}
& \includegraphics[width = 0.15\textwidth]{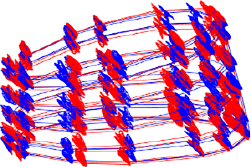} 
& \includegraphics[width = 0.15\textwidth]{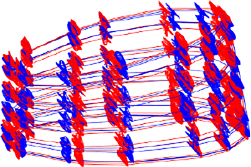}
& \includegraphics[width = 0.15\textwidth]{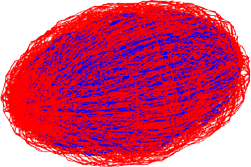}
\\
$\pm9^{\circ}$
& \includegraphics[width = 0.15\textwidth]{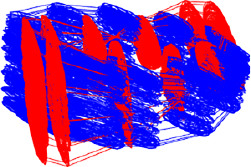}
& \includegraphics[width = 0.15\textwidth]{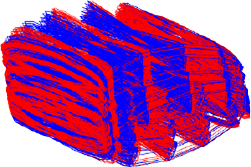} 
& \includegraphics[width = 0.15\textwidth]{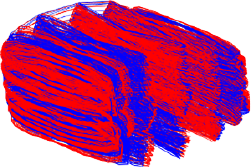}
& \includegraphics[width = 0.15\textwidth]{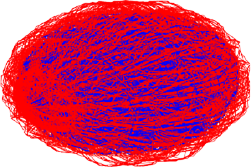}
\\
$\pm10^{\circ}$
& \includegraphics[width = 0.15\textwidth]{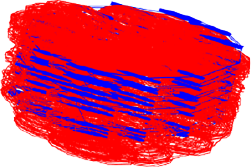}
& \includegraphics[width = 0.15\textwidth]{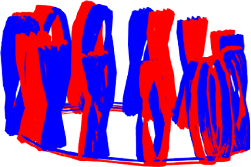} 
& \includegraphics[width = 0.15\textwidth]{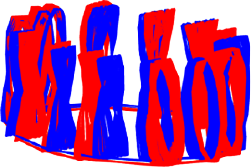}
& \includegraphics[width = 0.15\textwidth]{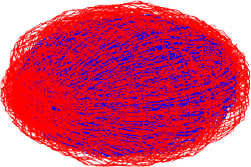}
\\\hline\noalign{\smallskip}
\end{tabularx}

	\vskip -0.5cm
	\caption{Feature-level alignment of manifolds under the influence of different levels of joint angle noise: Each graph visualises a different way of aligning the two pendulum manifolds. Each row shows the results for a different level of joint angle noise.}
	\label{fig:5} 
\end{figure}

\begin{figure}[!ht]
		\small
\begin{tabularx}{1\textwidth}{M{0.06\textwidth}|M{0.15\textwidth}M{0.15\textwidth}M{0.15\textwidth}M{0.15\textwidth}}
\hline\noalign{\smallskip}
\multirow{2}{*}{\parbox{1.0cm}{\centering Noise\\level}} 
& 
\multicolumn{1}{c}{\multirow{2}{*}{\parbox{2.4cm}{\centering Method 1: Pro-  crustes analysis}}} 
&
\multicolumn{1}{c}{\multirow{2}{*}{\parbox{2.4cm}{\centering Method 2: Local  Laplacian}}} 
&
\multicolumn{1}{c}{\multirow{2}{*}{\parbox{2.4cm}{\centering Method 3: Local weights}}}
&
\multicolumn{1}{c}{\multirow{2}{*}{\parbox{2.4cm}{\centering Method 4: Global distance}}}
\\ & & & & \\
\hline\\
$\pm0^{\circ}$
& \includegraphics[width = 0.15\textwidth]{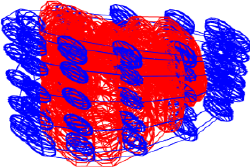}
& \includegraphics[width = 0.15\textwidth]{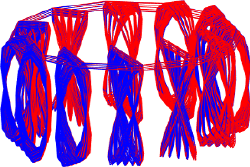} 
& \includegraphics[width = 0.15\textwidth]{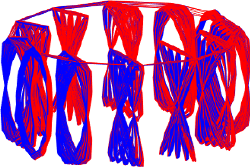}
& \includegraphics[width = 0.15\textwidth]{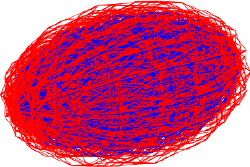}
\\
$\pm1^{\circ}$
& \includegraphics[width = 0.15\textwidth]{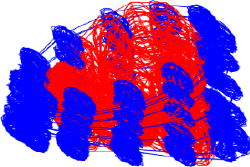}
& \includegraphics[width = 0.15\textwidth]{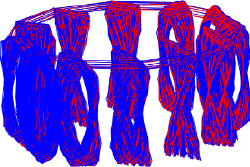} 
& \includegraphics[width = 0.15\textwidth]{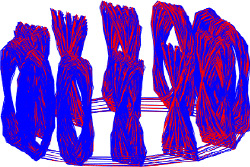}
& \includegraphics[width = 0.15\textwidth]{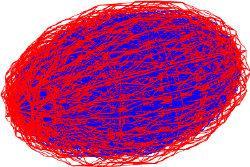}
\\
$\pm2^{\circ}$
& \includegraphics[width = 0.15\textwidth]{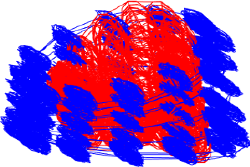}
& \includegraphics[width = 0.15\textwidth]{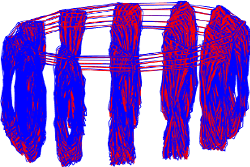} 
& \includegraphics[width = 0.15\textwidth]{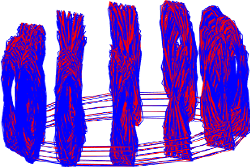}
& \includegraphics[width = 0.15\textwidth]{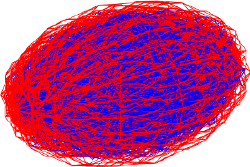}
\\
$\pm3^{\circ}$
& \includegraphics[width = 0.15\textwidth]{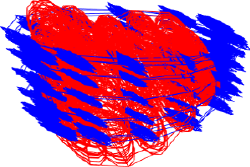}
& \includegraphics[width = 0.15\textwidth]{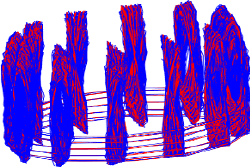} 
& \includegraphics[width = 0.15\textwidth]{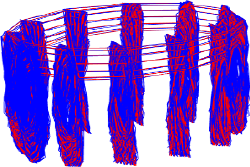}
& \includegraphics[width = 0.15\textwidth]{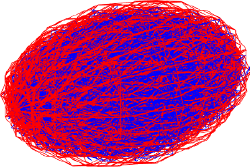}
\\
$\pm4^{\circ}$
& \includegraphics[width = 0.15\textwidth]{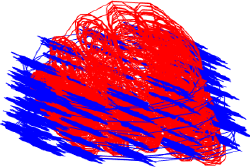}
& \includegraphics[width = 0.15\textwidth]{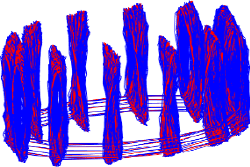} 
& \includegraphics[width = 0.15\textwidth]{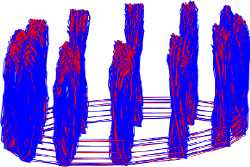}
& \includegraphics[width = 0.15\textwidth]{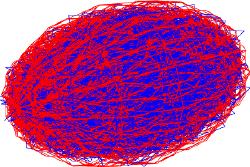}
\\
$\pm5^{\circ}$
& \includegraphics[width = 0.15\textwidth]{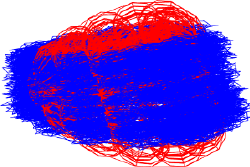}
& \includegraphics[width = 0.15\textwidth]{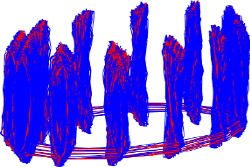} 
& \includegraphics[width = 0.15\textwidth]{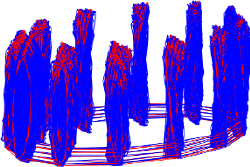}
& \includegraphics[width = 0.15\textwidth]{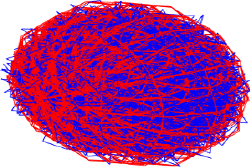}
\\
$\pm6^{\circ}$
& \includegraphics[width = 0.15\textwidth]{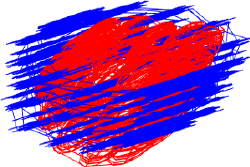}
& \includegraphics[width = 0.15\textwidth]{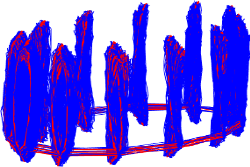} 
& \includegraphics[width = 0.15\textwidth]{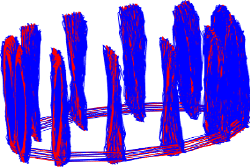}
& \includegraphics[width = 0.15\textwidth]{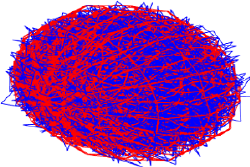}
\\
$\pm7^{\circ}$
& \includegraphics[width = 0.15\textwidth]{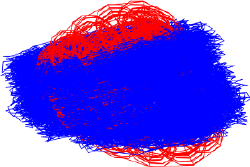}
& \includegraphics[width = 0.15\textwidth]{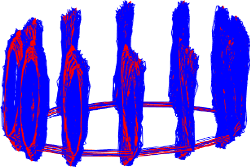} 
& \includegraphics[width = 0.15\textwidth]{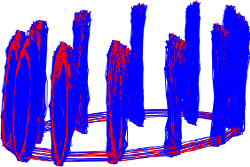}
& \includegraphics[width = 0.15\textwidth]{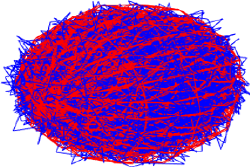}
\\
$\pm8^{\circ}$
& \includegraphics[width = 0.15\textwidth]{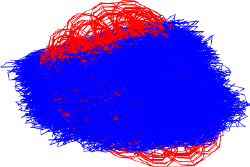}
& \includegraphics[width = 0.15\textwidth]{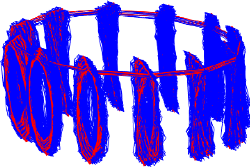} 
& \includegraphics[width = 0.15\textwidth]{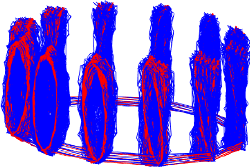}
& \includegraphics[width = 0.15\textwidth]{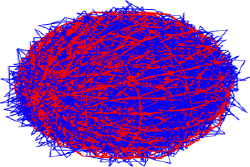}
\\
$\pm9^{\circ}$
& \includegraphics[width = 0.15\textwidth]{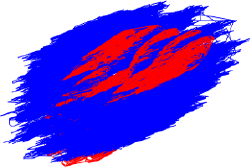}
& \includegraphics[width = 0.15\textwidth]{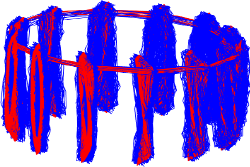} 
& \includegraphics[width = 0.15\textwidth]{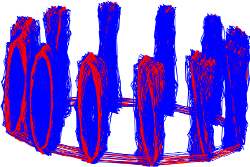}
& \includegraphics[width = 0.15\textwidth]{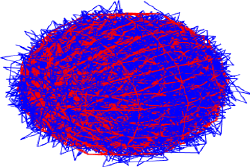}
\\
$\pm10^{\circ}$
& \includegraphics[width = 0.15\textwidth]{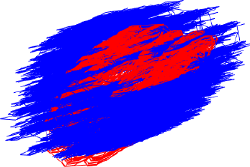}
& \includegraphics[width = 0.15\textwidth]{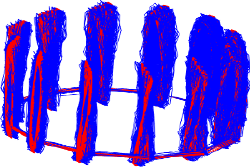} 
& \includegraphics[width = 0.15\textwidth]{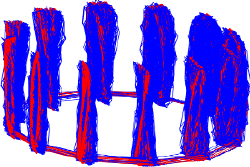}
& \includegraphics[width = 0.15\textwidth]{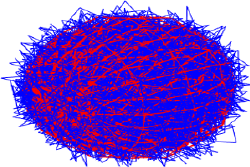}
\\\hline\noalign{\smallskip}
\end{tabularx}

	\vskip -0.5cm
    \caption{Feature-level alignment of manifolds under the influence of different levels of coordinate noise: Each graph visualises a different way of aligning the two pendulum manifolds. Each row shows the results for a different level of coordinate noise.}
	\label{fig:6} 
\end{figure}

\begin{figure}[!ht]
  \centering     
  \leavevmode
  \subfigure[]{\includegraphics[width=0.86\textwidth]{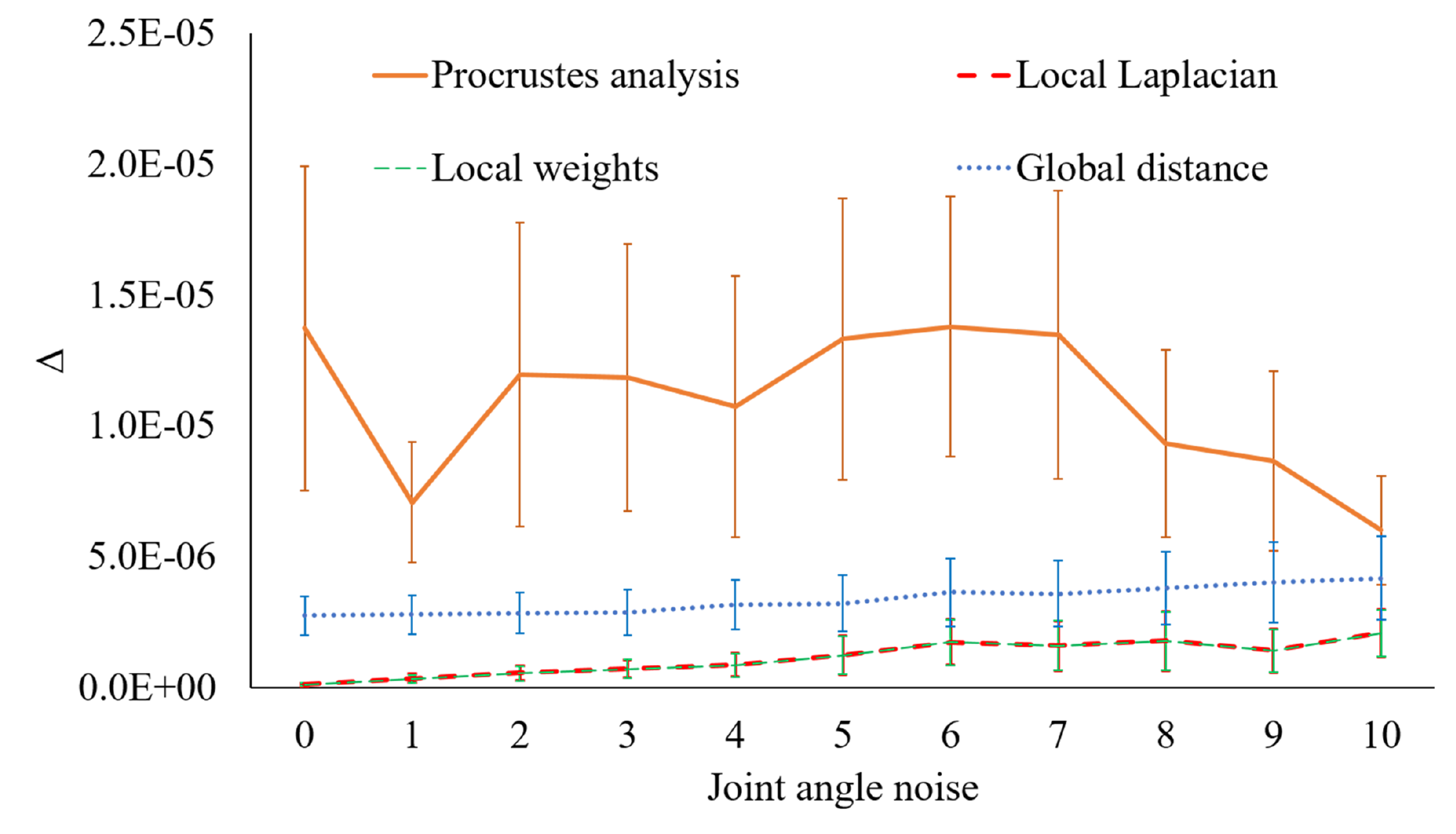}}\\
  \subfigure[]{\includegraphics[width=0.86\textwidth]{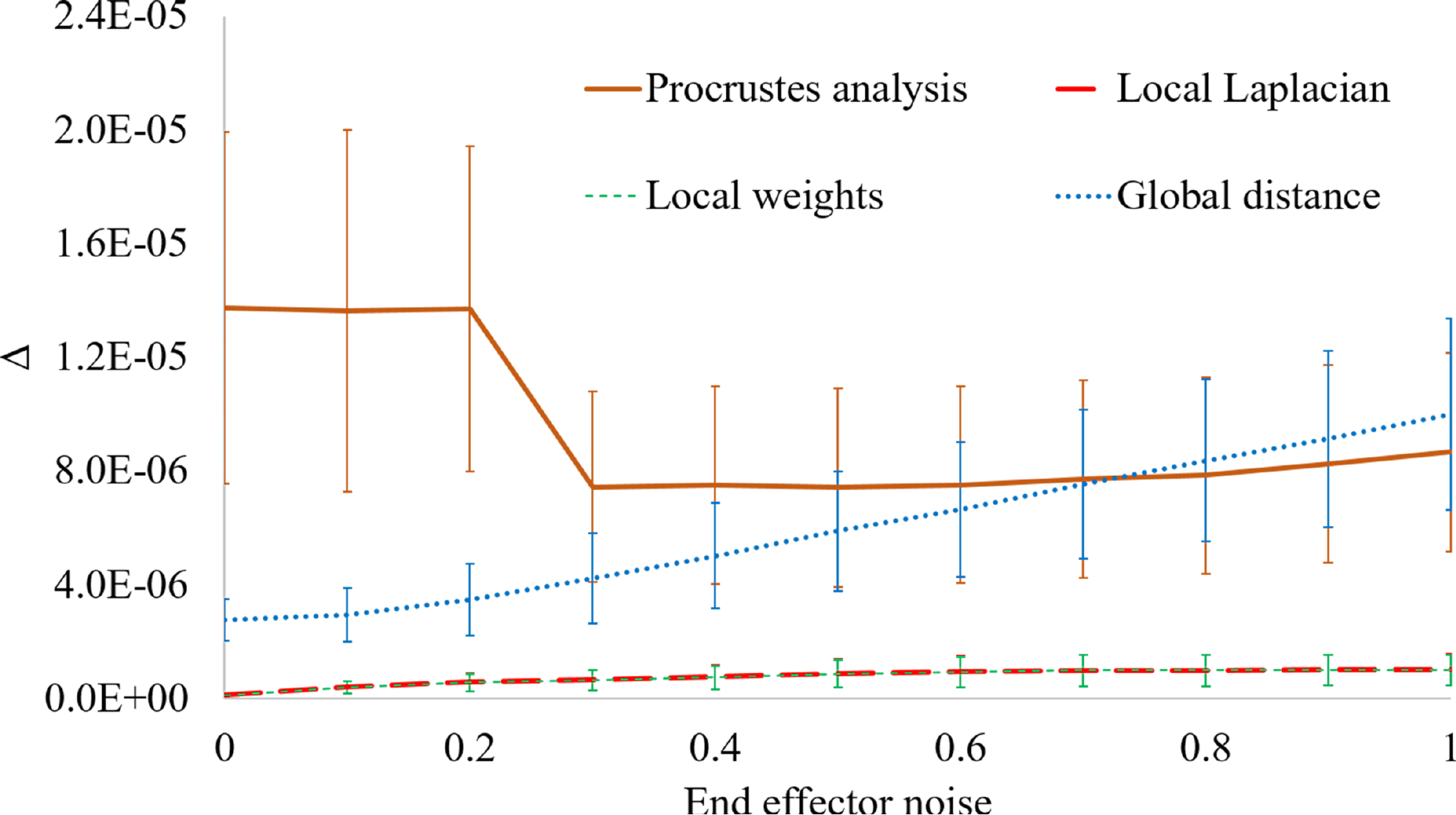}}
  \vskip 0.2in
  \caption{Performance graphs of manifold alignment methods that align the pendulum data with additional (a) joint angle noise and (b) end effector noise. The $x$-axes show the noise range. The $y$-axes show the averages $\Delta$ where the error bars are the standard deviations $\sigma$ of the correspondence distances of the aligned manifolds. $\Delta$ and $\sigma$ for  manifold alignment using local weights and local Laplacian are lower than for the other methods and this result is not much affected by any of the added noise.}
  \label{fig:7}
\end{figure}

Table~\ref{table:1} shows results for the feature-level cases where  $\Delta$ and $\sigma$ were lowest for Method 3. The execution time of each method for same datasets is shown in the last column of~\ref{table:1}. Due to the involvement of high dimensional matrix multiplication, the manifold alignment using global distance required a significantly higher execution time than the others.
\begin{table}[!ht]
\caption{Summary of the performance of manifold alignment methods: Feature-level manifold alignment using local weights achieved the lowest $\Delta$ and $\sigma$ (3rd row).}
\centering
\begin{tabularx}{\textwidth}{c @{\extracolsep{\fill}} lllll}
\hline\noalign{\smallskip}
\multicolumn{1}{c}{}&\multicolumn{1}{c}{\textbf{Method}}&\multicolumn{1}{l}{\textbf{Level}}&\multicolumn{1}{c}{\textbf{$\Delta$}}&\multicolumn{1}{l}{\textbf{$\sigma$}}&\multicolumn{1}{l}{\textbf{Time(s)}}\\
\noalign{\smallskip}\hline
\noalign{\smallskip}
1 & Procrustes  analysis & Feature   & $1.37\times10^{-05}$  & $6.20\times10^{-06}$ & 393\\
2 & Local Laplacian    & Feature  	& $1.22\times10^{-07}$ & $5.08\times10^{-08}$ & 382 \\
3 & {\bf Local weights}  & {\bf Feature} 	 & $\boldsymbol{1.22\times10^{-07}}$  & $\boldsymbol{4.92\times10^{-08}}$ & {\bf 343}\\
4 & Global distance  & Feature   & $2.76\times10^{-06}$ & $7.38\times10^{-07}$  & $6.3\times10^4$\\
\hline
\end{tabularx}

\label{table:1}
\end{table}
Figures~\ref{fig:5} and ~\ref{fig:6} show the aligned manifolds in the feature level cases after application of joint angle and coordinate noise, respectively. Both limbs of the double pendulums were rotating in three-dimensional spherical motion. Ergo the overall motion manifold can be described as the cross product of two spheres (as indicated in column 7 of Figure~\ref{fig:4}). The visualisations of the outcome of manifold alignment preserving global geometry are shown in column 4 of Figures~\ref{fig:5} and~\ref{fig:6}. The results seemed to be robust to noise and the visualisation were as expected. On the other hand, manifold alignments using the local Laplacian and local weights resulted in lower $\Delta$ and $\sigma$ (Figure~\ref{fig:7}) but visually more complex outcomes that also significantly varied under the influence of noise (Figures~\ref{fig:5} and ~\ref{fig:6}).

\section{Discussion and Conclusion}\label{sec:Summary and Discussion}
This study compared four different manifold alignment methods (Figure~\ref{fig:3}). The data sets were generated by simulating the motion of two three-dimensional double pendulums that differed in the lengths of their arms. Where the joint angles of the two pendulums were equal, they were recorded as corresponding pairs, and this was exploited in the manifold alignment and performance evaluation. 

We also investigated the effects of two different types of noise on the alignment of these data sets. First, random noise was added to the joint angles to observe the stability of the alignment with respect to actuator irregularities. Then, random noise was added to the end effector coordinates to observe the stability of the alignment methods with respect to jittery motions of the arm or noise in the data recordings. The performance of these methods was evaluated quantitatively by measuring the proximity of the corresponding points and qualitatively using visualisations. The experiments were conducted in Matlab 2016b on a high-performance computer cluster equipped with Xenon CPUs and 500GB RAM where all experiments for this paper could be executed in a total running time of about eighteen days. Individual running times are reported in Table~\ref{table:1}.

The approach of alignment preserving global geometry produced the most convincing visualisations but was also much slower than the other methods.  The methods using local weights resulted in the numerically smallest alignment errors (Table~\ref{table:1}). One possible interpretation is that local methods had an advantage because the experiments were limited to data from two similarly configured pendulums. Another interpretation is that parts of the manifolds collapsed, i.e., they were projected onto line segments during the manifold learning process. This is supported by the visualisations in Figures~\ref{fig:4},~\ref{fig:5} and~\ref{fig:6}. The observed instabilities of the local and instance level methods require further investigation.

Traditional robotic motion control engines often rely on inverse kinematics and can be difficult to adapt when the robot configuration changes~\cite{mosavi:2017}. Chalodhorn and Rao found that direct use of kinematics data from a human motion capture system to replicate human movement in a robot can result in dynamically unstable motion~\cite{Chalodhorn:2010}. This instability is the result of differences in the degrees-of-freedom and inadequate physical parameters for a robot to match human motion. The outcomes of the present study support the hypothesis that manifold alignment could provide a mapping between motion trajectories of similar manifolds. Moreover, a feature-level alignment approach could align trajectories on manifolds that may include out-of-sample extensions that may be generated during changes of configurations.

One of the purposes of dimensionality reduction is to process data faster while requiring less memory  than the calculations in high dimensions. Feature-level alignment provides a function that can map instances from high-dimensional feature space to low-dimensional alignment space. This function can map any new sample, whether it was included in the training data set or not, to the aligned manifolds. In contrast, the instance-level methods require a complete recalculation of the alignment mapping for each new sample. Therefore, feature-level alignment can perform better than instance-level alignment when out-of-sample extension processing is necessary~\cite{wang:2010}. The latter could be the case, for example, when the robot configuration changes.

In future research, the findings of this study could possibly become useful for the task of motion imitation~\cite{pan:2010}. Using manifold alignment transfer learning between robot arms could be achieved by copying previously trained stable motions from one robot to another where the robots' arms could have different limb length proportions. 

\section*{Acknowledgements}
FA was supported by by a UNRSC50:50 PhD scholarship at the University of Newcastle, Australia. The authors are grateful to the UON ARCS team who facilitated access to the UON high performance computing system.

\bibliographystyle{hplain}
\bibliography{bibliography} 

\end{document}